\documentclass[twocolumn]{el-author}

\newcommand{\ua}{\uparrow}
\newcommand{\nc}{\newcommand}
\nc{\da}{\downarrow} \nc{\hc}{\hat{c}} \nc{\hS}{\hat{S}}
\nc{\bra}{\langle} \nc{\ket}{\rangle} \nc{\eq}{equation (\ref}
\nc{\h}{\hat} \nc{\hT}{\h{T}}\nc{\be}{\begin{eqnarray}}
\nc{\ee}{\end{eqnarray}}\nc{\rd}{\textrm{d}}\nc{\e}{eqnarray}\nc{\hR}{\hat{R}}\nc{\Tr}{\mathrm{Tr}}
\nc{\tS}{\tilde{S}}\nc{\tr}{\mathrm{tr}}\nc{\8}{\infty}\nc{\lgs}{\bra\ua,\phi|}\nc{\rgs}{|\ua,\phi\ket}
\nc{\hU}{\hat{U}}\nc{\lfs}{\bra\phi|}\nc{\rfs}{|\phi\ket}\nc{\hZ}{\hat{Z}}\nc{\hd}{\hat{d}}\nc{\mD}{\mathcal{D}}
\nc{\bd}{\bar{d}}\nc{\bc}{\bar{c}}\nc{\mc}{\mathcal}\nc{\ea}{eqnarray}\nc{\mG}{\mathcal{G}}\nc{\bce}{\begin{center}}
\nc{\ece}{\end{center}}
\date{}

\begin{document}

\title{An accelerated CLPSO algorithm}

\author{M. O. Bin Saeed, M. S. Sohail, S. Z. Rizvi, M. Shoaib and A. U. H. Sheikh}

\abstract{The particle swarm approach provides a low complexity solution to the optimization problem among various existing heuristic algorithms. Recent advances in the algorithm resulted in improved performance at the cost of increased computational complexity, which is undesirable. Literature shows that the particle swarm optimization algorithm based on comprehensive learning provides the best complexity-performance trade-off. We show how to reduce the complexity of this algorithm further, with a slight but acceptable performance loss. This enhancement allows the application of the algorithm in time critical applications, such as, real-time tracking, equalization etc.  }

\maketitle

\section{Introduction}
\label{intro}

The particle swarm optimization (PSO) algorithm was introduced in 1995 by Kennedy and Eberhart \cite{Eberhart95,Kennedy95}. It is a population based optimization algorithm, emulating swarm behavior that is observed in a herd of animals, a flock of birds or a school of fish. For function optimization purposes, the swarm consists of particles, hence the name particle swarm optimization. Each particle searches for a potential solution in a multi dimensional search space. The aim of the swarm is to converge to an optimum solution, global to the whole swarm. The estimate for each particle is tested using a certain fitness value that defines the goodness of the estimate. Every particle combines its own best attained solution with the best solution of the whole swarm to adapt its search pattern.

Several variants have been suggested in the literature to improve the performance of the PSO algorithm. Each variant has its own advantages and disadvantages. Among the most popular variants, the comprehensive learning PSO (CLPSO) and orthogonal learning PSO (OLPSO) algorithms provide the best performance over a wide range of test functions \cite{Liang06,Zhan11}. Results in \cite{Liang06} and \cite{Zhan11} show the superiority of these two algorithms over the other variants. However, both these algorithms are computationally very complex. In general, the CLPSO variant provides the best performance-complexity trade-off among all other existing PSO variants. Still, it is too complex for many applications, such as channel equalization and radar detection etc. It also tends to be slow converging as it attempts to provide highly accurate results. We note that most applications do not require such a high level of accuracy. Here, we propose a fast converging, low complexity solution that achieves reasonably good level of accuracy. The cost is a slight degradation in performance, which is still highly acceptable for the aforementioned applications.

\section{Comprehensive Learning Particle Swarm Optimizer (CLPSO)}
\label{CLPSO}

Consider a {\em D}-dimensional hyperspace in which a swarm of {\em N} particles is trying to find the optimum solution, given by the position vector ${\bf X}=\left[x^1,x^2,\cdots,x^D\right]^T$. The individual position of each particle $k$ is denoted by the {\em D}-dimensional vector ${\bf X}_k=\left[x_{k}^1,x_{k}^{2},\cdots,x_{k}^{D}\right]^T$. Similarly, the velocity vector of each particle $k$ is given by the vector ${\bf V}_k=\left[v_{k}^{1},v_{k}^{2},\cdots,v_{k}^{D}\right]^T$. Each particle updates its position based on its own best recorded position as well as the best recorded position of the whole swarm. Let the best individual position vector of each particle and the best global position vector be denoted by the vectors ${\bf P}_k=\left[p_{k}^{1},p_{k}^{2},\cdots,p_{k}^{D}\right]^T$ and ${\bf G}_k=\left[g^{1},g^{2},\cdots,g^{D}\right]^T$, respectively. Then the velocity update equation for the PSO algorithm is given by:
\begin{align}\label{v_orig}
v_{k}^{d} \left(i+1\right) = v_{k}^{d} \left(i\right) + c_1.r_1.\left(p_{k}^{d} - x_{k}^{d} \left(i\right) \right) + c_2.r_2.\left(g^{d} - x_{k}^{d} \left(i\right) \right),
\end{align}
where $k = 1,\cdots,N$ is the particle index, $d = 1,\cdots,D$ is the dimension index, $i$ denotes the time index, $c_1$ and $c_2$ are positive constants known as the acceleration coefficients, and $r_1$ and $r_2$ are uniformly distributed random numbers within the range $\left[0,1\right]$. The global best is generated from the neighborhood of each particle. In general, the velocity and the position for each particle are bounded within a predefined limit, $\left[-V_{max},V_{max}\right]$ and $\left[X_{min},X_{max}\right]$, respectively. The bounds ensure that the particles do not diverge from the search hyperspace.

The CLPSO algorithm divides the unknown estimate into sub-vectors \cite{Liang06}. For each sub-vector, a particle chooses two random neighbor particles. The neighbor particle that gives the best fitness value for that particular sub-vector is chosen as an exemplar. The combined result of all sub-vectors gives the overall best vector, which is then used to perform the update. If the particle stops improving for a certain number of iterations then the neighbor particles for the sub-vectors are changed. The velocity update equation for the CLPSO algorithm is given by \cite{Liang06}:
\begin{align}
v_{k}^{d} \left(i+1\right) =& w_k(i).v_{k}^{d} \left(i\right) + c_k.r_k.\left({\hat p}_{k}^{d} - x_{k}^{d} \left(i\right) \right),
\end{align}
where $w_k(i)$ is a time-varying weighting coefficient, $\hat p_k^d$ gives the overall best position value for dimension $d$ of particle $k$, $c_k$ is the acceleration coefficient and $r_k$ is a uniformly distributed random number. The overall CLPSO algorithm is thus defined by the following set of equations \cite{Liang06}:
\begin{align}
{\bf V}_{k} \left(i+1\right) =& w_k (i).{\bf V}_{k} \left(i\right) + c_k.r_k.\left({\bf \hat P}_{k} - {\bf X}_{k} \left(i\right) \right) \label{v1} \\
{\bf X}_{k} \left(i+1\right) =& {\bf X}_{k} \left(i\right) + {\bf V}_{k} \left(i+1\right),
\end{align}

\section{The Accelerated CLPSO Algorithm}
\label{ACLPSO}

An event-triggering approach is applied to the CLPSO algorithm. The accelerating coefficient $c_k$ is set to $0$ if the distance value $\left|{\hat p}_{k}^{d} - x_{k}^{d} \left(i\right)\right| \le \gamma$, where $\gamma$ is a certain threshold. This ensures that the particle stays close to the best obtained vector. However, each dimension is to be treated separately and the accelerating coefficient $c_k$ is converted into a vector where each individual value $c_k^d$ is treated based on whether the distance value for that dimension is within the threshold or not. Thus, the accelerating coefficient is governed by the following equation:
\begin{align}
c_k^d\left( i \right) = \left\{ \begin{array}{l}
0,\hspace{0.4cm}{\rm{if }}\left| {\hat p_k^d - x_k^d\left( i \right)} \right| \le \gamma \\
{\mathcal C}\hspace{0.5cm}{\rm{otherwise}}
\end{array} \right.
\end{align}

The above formulation stems from two observations. One observation is that all dimensions do not require an update at every iteration. Secondly, every iterative step does not contribute significantly towards an improved update. The update equations are executed every time, resulting in extra computations. Our proposed step aims to remove the insignificant update steps, resulting in a faster algorithm. We term our proposed algorithm as an accelerated CLPSO algorithm (ACLPSO).

\section{Test Functions}
\label{test_func}

Several test functions are available in the literature that can be used to test the performance of the different variants of the PSO algorithms \cite{Liang06,Zhan11}. The algorithms aim to minimize the fitness value for these test functions. Due to the paucity of space, we choose only the five functions given in Table \ref{Table0}:
\begin{table}[h]
\processtable{Test functions and their equations.}
{\begin{tabular}{c l l}
\hline
&&\\
 & {\bf Function names} & {\bf Function equations}\\
\hline
1. & {\em Sphere} & $f_1 (x) = \sum\limits_{i = 1}^D {x_i^2}$\\
2. & {\em Rosenbrock} & $f_2 (x) = \sum\limits_{i = 1}^D {\left( {100{{\left( {x_i^2 - {x_{i + 1}}} \right)}^2} + {{\left( {xi - 1} \right)}^2}} \right)}$\\
3. & {\em Rastrigin} & $f_3 (x) = \sum\limits_{i = 1}^D {\left(x_i^2 - 10{\rm cos}\left(2\pi x_i\right) + 10\right)}$\\
4. & {\em Griewank} & $f_4 (x) = \sum\limits_{i = 1}^D {\frac{x_i^2}{4000}} - \prod\limits_{i = 1}^D {{\rm cos} \left( \frac{x_i}{\sqrt{i}} \right) + 1}$\\
5. & {\em Ackley} & $f_5 (x) = -20 {\rm exp} \left(-0.2\sqrt{\frac{1}{D}\sum\limits_{i = 1}^D {x_i^2}}\right)$\\
 &&\hspace{1cm}$- {\rm exp} \left( \frac{1}{D} \sum\limits_{i = 1}^D {{\rm cos}\left(2\pi x_i\right)}\right) + 20 + {\rm e}$\\
\hline
\label{Table0}
\end{tabular}}{}
\end{table}

\section{Results and Discussion}
\label{ACLPSO_perf}

In this section, we compare the performance of the standard CLPSO algorithm with the newly proposed ACLPSO algorithm using the test functions defined above. The swarm has 40 particles, with each particle having 30 dimensions. All results are averaged over 200 experiments.

Table \ref{Table1} gives the results of the simulations. Two different values of $\gamma$ are chosen for detailed analysis. The table shows the mean fitness values for the CLPSO and ACLPSO algorithms when they run for the complete 5000 iterations. As can be seen from (\ref{v1}), there are a total of 3 multiplications for updating each dimension per particle. The first multiplication occurs every time for both the algorithms. The remaining two multiplications occur in the proposed algorithm only if the threshold value is exceeded whereas they always occur for the CLPSO algorithm. As a result, the number of computations is reduced. The $\%$-age computations column lists the amount of computations the proposed algorithm requires relative to the standard CLPSO algorithm. The effective iterations column gives the equivalent iteration number at which the CLPSO algorithm has the same number of computations as the proposed algorithm. The effective mean is the fitness value of the CLPSO algorithm after the number of effective iterations. As can be seen, the proposed algorithm provides a very fast and low-complexity result for all test functions. The only trade-off is that the proposed algorithm does not achieve the same final value as the CLPSO algorithm. However, this trade-off is acceptable as the applications that require fast convergence, generally, do not require accuracy of more than a few decimal places. This is the main focus of our work and, as can be seen clearly, the outcome is more than satisfactory.
\begin{table}[h]
\processtable{Simulation results.}
{\begin{tabular}{|cc|c|ccc|c|}
\hline
&&&&&&\\
{\bf Func.} & ${\bf \gamma}$ & {\bf CLPSO} & & {\bf ACLPSO} & & {\bf CLPSO} \\
 & & Actual & Mean & $\%$-age & Eff & Eff \\
 & & Mean & & Comp & Iter & Mean \\
\hline
$f_1 (x)$ & $1e-1$ & $7e-14$ & $6e-2$ & $43.34 \%$ & 2167 & 0.47\\
 & $1e-3$ & $7e-14$ & $6e-6$ & $67.33 \%$ & 3367 & $1e-2$\\
\hline
$f_2 (x)$ & $1e-1$ & $28.17$ & $34.24$ & $44.31 \%$ & 2216 & 68.56\\
 & $1e-3$ & $28.17$ & $28.23$ & $67.78 \%$ & 3390 & $29.44$\\
\hline
$f_3 (x)$ & $1e-1$ & $2e-11$ & $11.87$ & $43.57 \%$ & 2179 & 57.42\\
 & $1e-3$ & $2e-11$ & $1e-3$ & $67.33 \%$ & 3367 & $2.79$\\
\hline
$f_4 (x)$ & $1e-1$ & $9e-15$ & $3e-3$ & $43.24 \%$ & 2162 & $2e-2$\\
 & $1e-3$ & $9e-15$ & $3e-7$ & $67.29 \%$ & 3365 & $6e-4$\\
\hline
$f_5 (x)$ & $1e-1$ & $2e-7$ & $0.28$ & $43.35 \%$ & 2168 & 1.04\\
 & $1e-3$ & $2e-7$ & $2e-3$ & $67.34 \%$ & 3367 & 0.10\\
\hline
\label{Table1}
\end{tabular}}{}
\end{table}

\begin{figure}[h]
\centering{\includegraphics[width=80mm]{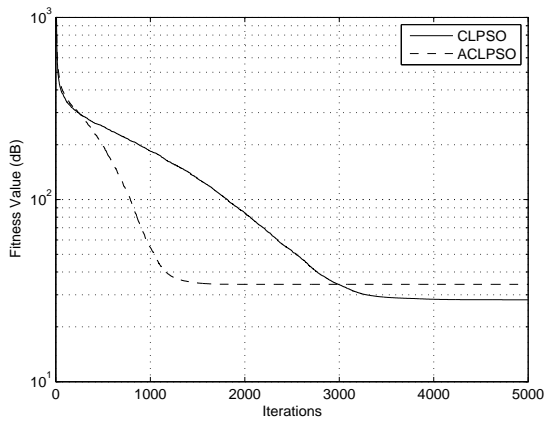}}
\caption{Performance comparison for Rosenbrock function with $\gamma = 1e-1$.
\source{}}\label{ros_acl}
\end{figure}

\begin{figure}[h]
\centering{\includegraphics[width=80mm]{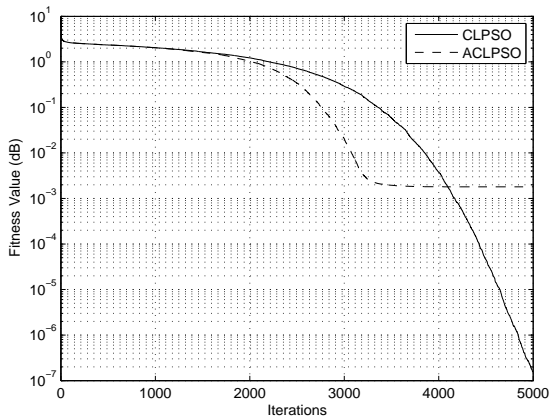}}
\caption{Performance comparison for Ackley function with $\gamma = 1e-3$.
\source{}}\label{ack_acl}
\end{figure}

The simulation plots are shown in Figs. \ref{ros_acl} and \ref{ack_acl} for only two of the results. These results corroborate the results in the table and the discussion about the results.


\section{Conclusion}
\label{conc}

We have proposed an accelerated comprehensive learning PSO algorithm that provides a fast and low complexity solution for time critical applications. The simulation results show that the proposed algorithm performs faster than the standard CLPSO algorithm with reduced computations and an acceptable degradation in performance.

\vskip3pt

\vskip5pt

\noindent M. O. Bin Saeed, M. S. Sohail and A. U. H. Sheikh (\textit{Department of Electrical Engineering, College of Engineering Sciences, King Fahd University of Petroleum $\&$ Minerals, Dhahran 31261, Saudi Arabia})\\
\vskip3pt

\noindent E-mail: mobs@kfupm.edu.sa \\
\vskip5pt
\noindent S. Z. Rizvi (\textit{College of Engineering, University of Georgia, Athens, GA 30602, USA}) \\
\vskip5pt

\noindent M. Shoaib (\textit{Prince Sultan Advanced Technologies Research Institute, College of Engineering, King Saud University, Riyadh 11421, Saudi Arabia}) \\ \\

\end{document}